\documentclass{article} 
\usepackage[preprint]{colm2025_conference}

\usepackage{microtype}
\usepackage{hyperref}
\usepackage{url}
\usepackage{booktabs}
\usepackage{graphicx}
\usepackage{subcaption}
\usepackage{amsmath}
\usepackage{array}
\usepackage{wrapfig}
\usepackage{cleveref}

\usepackage{lineno}

\definecolor{darkblue}{rgb}{0, 0, 0.5}
\hypersetup{colorlinks=true, citecolor=darkblue, linkcolor=darkblue, urlcolor=darkblue}

\title{Analysis of Attention in Video Diffusion Transformers}

\author{
    \textbf{Yuxin Wen}$^{1}$\thanks{Correspondence to: ywen@umd.edu}, \quad \ 
    \textbf{Jim Wu}$^{2}$, \quad \ 
    \textbf{Ajay Jain}$^{2}$, \quad \ 
    \textbf{Tom Goldstein}$^{1}$, \quad \ 
    \textbf{Ashwinee Panda}$^{1}$ \\
    $^{1}$University of Maryland \quad \ 
    $^{2}$GenmoAI
}

\begin{document}

\ifcolmsubmission
\linenumbers
\fi

\maketitle

\newcommand{\vdit}{VDiT}
\newcommand{\dit}{DiT}

\begin{abstract}
We conduct an in-depth analysis of attention in video diffusion transformers (\vdit{}s) and report a number of novel findings. We identify three key properties of attention in \vdit{}s: \textbf{Structure}, \textbf{Sparsity}, and \textbf{Sinks}. \textbf{Structure:} We observe that attention patterns across different \vdit{}s exhibit similar structure across different prompts, and that we can make use of the similarity of attention patterns to unlock video editing via self-attention map transfer. \textbf{Sparse:} We study attention sparsity in \vdit{}s, finding that proposed sparsity methods do not work for all \vdit{}s, because some layers that are seemingly sparse cannot be sparsified. \textbf{Sinks:} We make the first study of attention sinks in \vdit{}s, comparing and contrasting them to attention sinks in language models. We propose a number of future directions that can make use of our insights to improve the efficiency-quality Pareto frontier for \vdit{}s. A web version with video examples is available \href{https://seasoned-draw-b97.notion.site/Analysis-of-Attention-in-Video-Diffusion-Transformers-1aea04ac6ca780c2b6b2cf6ed87e311f}{here}.
\end{abstract}

\section{Introduction}
Video diffusion models produce high-quality videos, but this comes at the cost of large and complex attention operations. Early video diffusion models \citep{ho2022video, li2022efficient} used U-Net architectures with factorized spatial and temporal attention. Recent progress has been driven by scaling model size of transformers on large-scale datasets \citep{sora2024, genmo2024mochi, kong2024hunyuanvideo, yang2024cogvideox, wan2025}. Modern Video Diffusion Transformers (\vdit{}s) use full bidirectional self-attention across a long context, treating space and time axes as a single flattened sequence. 
However, encoding just a few seconds of video requires tens to hundreds of thousands of tokens. With such long contexts, self-attention is the most computationally expensive component of transformers. Self-attention accounts for approximately $60\%$ of the total compute in Mochi-1~(10B parameters) \citep{genmo2024mochi}, with costs increasing quadratically with respect to height, width, and time.

The prominent role of attention in \vdit{}s motivates the study of the structure and patterns that emerge in these layers.  While this work is a fundamental study of these attention properties, we hope that an improved understanding of video attention will result in downtream advancements.  Indeed, this has been the case for 2D diffusions,
where attention manipulation has improved efficiency~\citep{yamaguchi2024exploring, pu2024efficient, yuan2024ditfastattn} and unlocked guided editing as a capability~\citep{hertz2022prompt, liu2024towards}.
 
In this work, we conduct an in-depth investigation into the characteristics of the attention module in \vdit{}s. We identify three key properties of attention in \vdit{}s: \textbf{Structure}, \textbf{sparsity}, and \textbf{sinks}. 
First, we examine the distribution of attention mass in video diffusion models, observing a strong and repeatable structure induced by patches attending to their spatial or temporal neighbors. We leverage this property to enable self-attention transfer: by transferring the attention structure from one prompt to a semantically similar prompt, we can impose the geometric structure induced by one prompt onto another. Furthermore, we identify specific layers that play distinct roles, such as controlling the camera angle.  

We then investigate the sparsity of attention matrices. We observe that sparsifying attention results in dramatic degradation in video quality. This, it turns out, is the result of a small number of layers that do not tolerate sparsity.  When these few layers are re-trained, we obtain a network that maintains strong performance even when 70\% sparsity is imposed on attention maps.
Additionally, we find that adjusting the temperature in \vdit{}s, which changes attention sparsity, leads to dramatic changes in the generated outputs. 

Finally, we identify the presence of attention sinks in \vdit{}s. Through our analysis, we show that attention sinks in \vdit{}s are more consistent and structured than those observed in language models. 

We also propose a simple mitigation strategy for non-sparse and attention-sink layers. By reinitializing and retraining these layers, we observe that all layers become sparsifiable and attention sinks are effectively eliminated.
Finally, we outline potential future research directions inspired by these findings.

\begin{figure}[!t]
    \centering
    \includegraphics[width=0.75\textwidth]{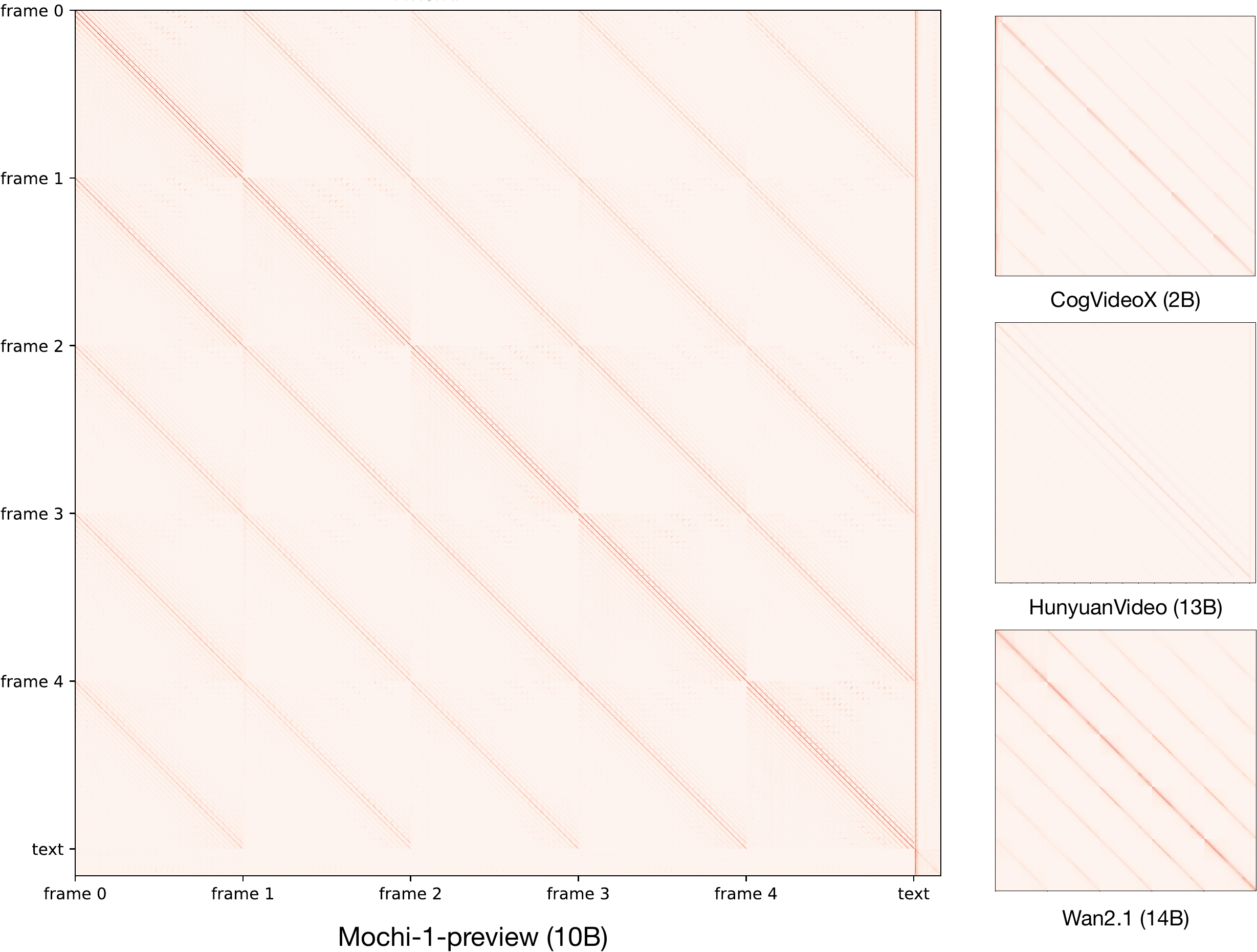}
    \caption{\textbf{Attention Maps.} Different models have the same structured attention patterns.}
    \label{fig:attention_map}
\end{figure}

\section{Background and Related Work}
Diffusion models \citep{ho2020denoising} generate data by reversing a gradual noising process, learning to iteratively denoise samples drawn from a known prior distribution (e.g., Gaussian noise). When applied to videos, a diffusion model generates coherent spatiotemporal sequences by progressively denoising noise into frames.

Early video diffusion models \citep{ho2022video, li2022efficient} used U-Net architectures with factorized spatial and temporal attention. Diffusion Transformers (\dit{}s) \citep{peebles2023scalable} are now popular due to their scalability. To improve efficiency, similar to latent image diffusion models \citep{rombach2022high}, \vdit{}s compress videos to a compact, continuous sequence with a VAE \citep{kingma2013auto}, and perform the diffusion process in latent space.

An input to a text-to-video diffusion model $X=[X_v, X_t]$ consists of vision tokens $V_X$, initially random Gaussian noise, and text tokens $X_t$, obtained from a text encoder. The vision tokens are flattened across temporal and spatial dimensions into $F \times S$ tokens, where $F$ is the number of latent frames, and $S$ is the number of tokens per frame. The concatenated tokens are fed into a Diffusion Transformer---similar to standard language model, but employing bidirectional rather than causal attention. Similar to mixture-of-expert models, some \vdit{}s~\citep{genmo2024mochi, kong2024hunyuanvideo} untie vision and text parameters to increase capacity and prevent interference between the modalities.

Understanding the behavior of self-attention can lead to more efficient and effective model designs. \citealt{child2019generating} discover the sparsity of language transformers and are able to perform more efficient attention with sparse factorizations. Meanwhile, for diffusion models, \citealt{hertz2023prompttoprompt} visualize and manipulate the cross-attention maps for image editing.

\section{Structured Attention}
The attention maps in~\Cref{fig:attention_map} are very similar, and for good reason: attention in \vdit{}s is \emph{structured} based on spatial-temporal locality. We see a clear diagonal stripe, and then off-diagonal stripes of varying strength. A diagonal stripe is spatial locality. Inside of one band on the diagonal, if we zoom in, we will see a visually distinct square for each frame. 
 
This is because pixels near each other exhibit spatial locality, with high attention weights. Off-diagonal stripes are temporal locality, and they decrease in strength as we move further off-diagonal because frames that are farther away from the current frame are less relevant.

\subsection{Self-Attention Transfer for Zero-Shot Video Editing}
\textit{Can a single attention map transfer to multiple prompts?
}We use Wan2.1-T2V-1.3B due to its architectural design, which applies self-attention over vision tokens before performing cross-attention with text tokens. This separation allows for more targeted attention transfer experiments, specifically on the self-attention layers operating over the visual input. During generation, given a source prompt, we directly take the attention map from the generation of the source prompt to the generation of the target prompt.

\begin{figure}[!ht]
    \centering
    \renewcommand{\arraystretch}{0.5}
    \begin{tabular}{@{}m{0.10\textwidth} m{0.85\textwidth}@{}}
        \centering \tiny (a): ``A car is driving on the highway.'' & \includegraphics[width=\linewidth]{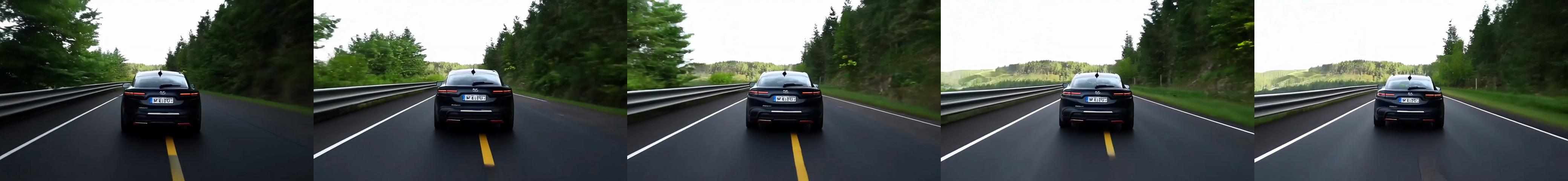} \\
        \centering \tiny (b): ``A dog is running on the grass.'' & \includegraphics[width=\linewidth]{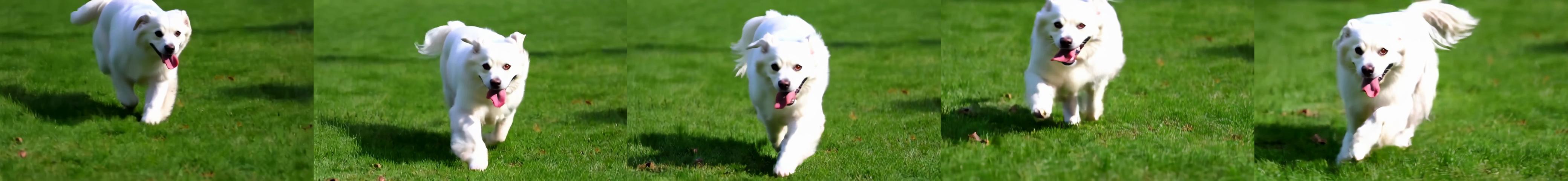} \\
        \centering \tiny Attention Map Transfer from (a) to (b)  & \includegraphics[width=\linewidth]{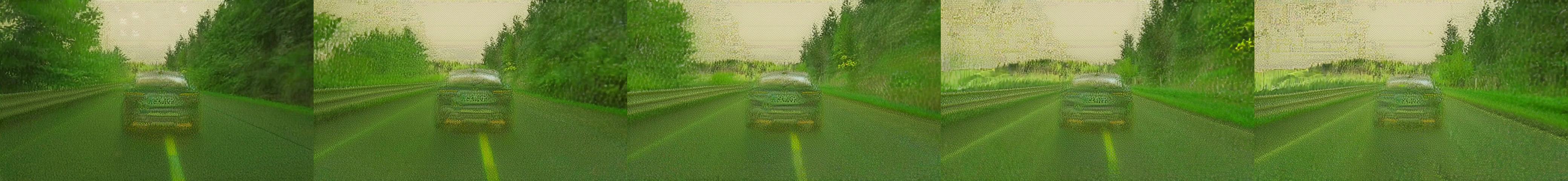} \\
    \end{tabular}
    \caption{\textbf{Attention Map Transfer to a Different Prompt.}}
    \label{fig:attention-transfer-dog}
\end{figure}

In~\Cref{fig:attention-transfer-dog}, we transfer the self-attention map from a car-driving prompt to a generation with a dog-running prompt. Surprisingly, this makes the video resemble the original car video—preserving its structure while ignoring the new prompt. This highlights how attention maps encode prompt-specific structure.

\begin{figure}[!ht]
    \centering
    \renewcommand{\arraystretch}{0.5}
    \begin{tabular}{@{}m{0.10\textwidth} m{0.85\textwidth}@{}}
        \centering \tiny (a): ``A car is driving on the highway.'' & \includegraphics[width=\linewidth]{assets/transfer/original_0.jpg} \\
        \centering \tiny (b): ``A red car is driving on the highway.'' & \includegraphics[width=\linewidth]{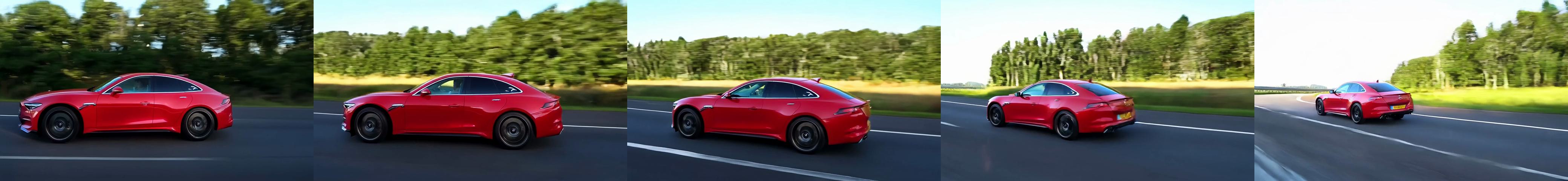} \\
        \centering \tiny Attention Map Transfer from (a) to (b)  & \includegraphics[width=\linewidth]{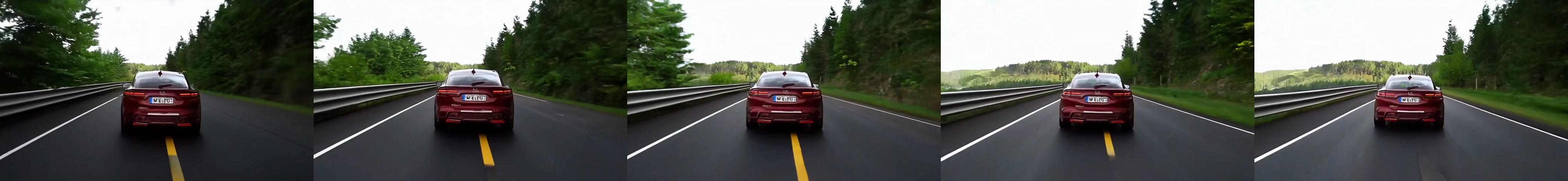} \\
    \end{tabular}
    \caption{\textbf{Attention Map Transfer to a Close Prompt.}}
    \label{fig:attention-transfer-red-car}
\end{figure}

Motivated by this, we explore the limits of fine-grained video editing. In~\Cref{fig:attention-transfer-red-car}, we conduct the same attention transfer experiment using a new prompt that is very similar to the original: \textit{``A car is driving on the highway''} vs. \textit{``A \textbf{red} car is driving on the highway.''}. As shown above, without attention transfer, the two generated videos differ significantly. Not just in the color of the car, but also in the perspective of the video and how it changes over time. From a user perspective, if we were happy with the cinematography of the original generated video and only wanted to change the color of the car, we would not be pleased that generating a video with the same seed leads to such a drastically different video. However, with attention map transfer, the resulting video is nearly identical to the original—except that the car is now red. This is a new capability for \emph{fine-grained video editing}, where only the attribute that the user wants to change is modified.

\begin{figure}[!ht]
\centering
    \renewcommand{\arraystretch}{0.5}
    \begin{tabular}{@{}m{0.10\textwidth} m{0.85\textwidth}@{}}
        \centering \tiny ``A car is driving on the highway in the winter.'' & \includegraphics[width=\linewidth]{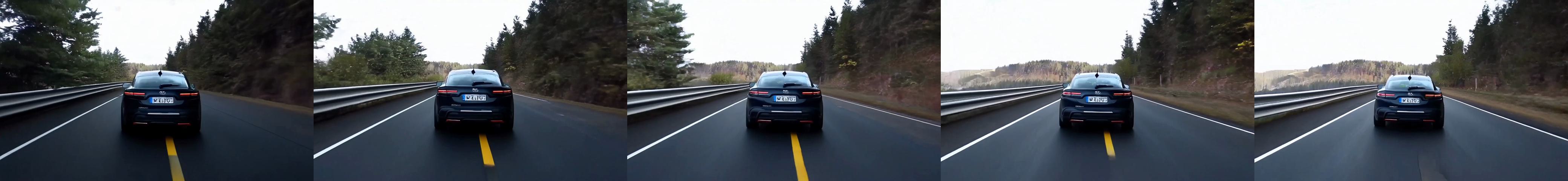} \\
        \centering \tiny ``A truck is driving on the highway.'' & \includegraphics[width=\linewidth]{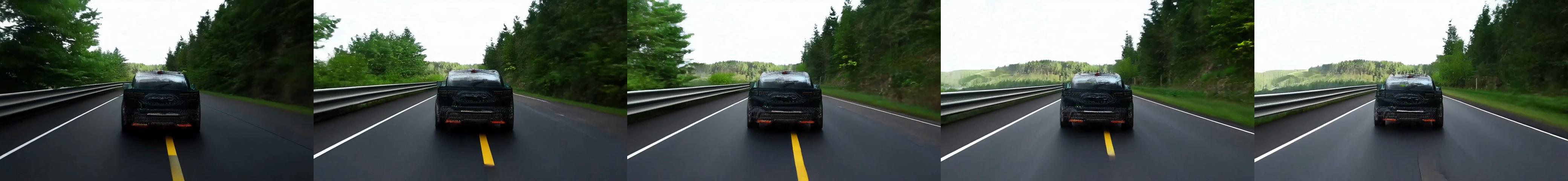} \\
    \end{tabular}
    \caption{\textbf{Attention Map Transfer to a Close Prompt.} The source prompt is ``A car is driving on the highway.'' from \Cref{fig:attention-transfer-red-car} (a).}
    \label{fig:attention-transfer-winter-truck}
\end{figure}

We can also try to change elements of the background or the car itself. In~\Cref{fig:attention-transfer-winter-truck}, we try to change the background to winter, and change the car to a truck. The transfer works well in the ``winter'' case. However, when the variation becomes larger—such as changing a car to a truck—the quality of the generation becomes limited.

\begin{figure}[!ht]
    \centering
    \renewcommand{\arraystretch}{0.5}
    \begin{tabular}{@{}m{0.10\textwidth} m{0.85\textwidth}@{}}
        \centering Layer $0$ & \includegraphics[width=\linewidth]{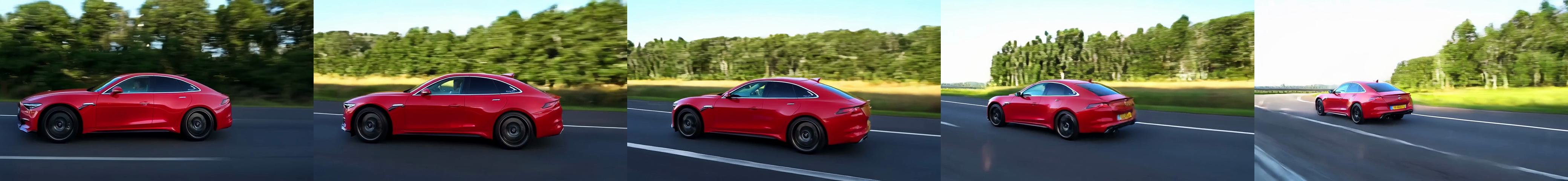} \\
        \centering Layer $3$ & \includegraphics[width=\linewidth]{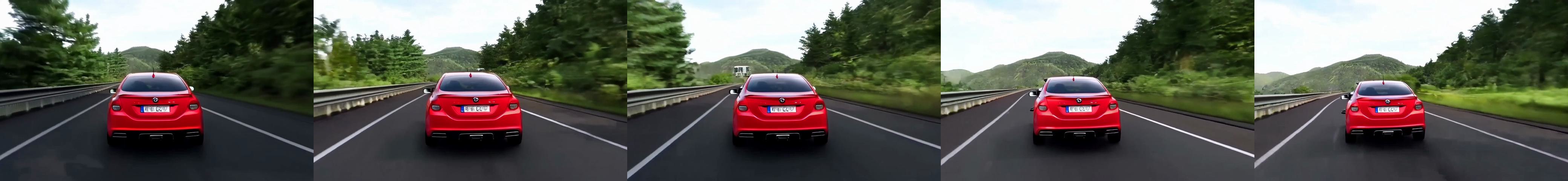} \\
        \centering Layer $19$ & \includegraphics[width=\linewidth]{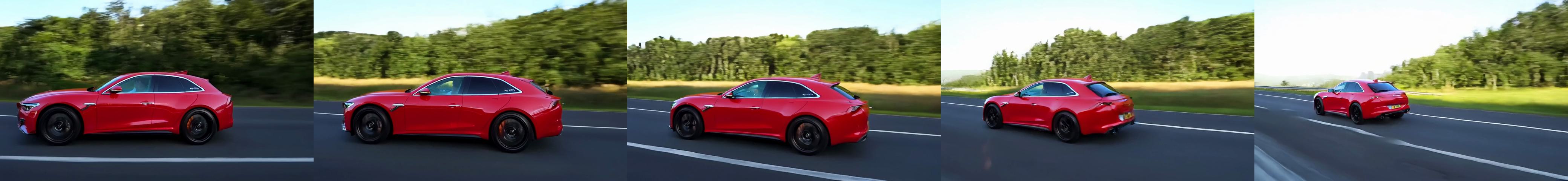} \\
    \end{tabular}
    \caption{\textbf{Attention Map Transfer to a Close Prompt only with One Layer.} The source prompt is ``A car is driving on the highway.'' from \Cref{fig:attention-transfer-red-car} (a).}
    \label{fig:attention-transfer-layerwise}
\end{figure}

In~\Cref{fig:attention-transfer-layerwise}, we performed attention transfer for each individual layer to identify which ones have the most significant impact. As shown below, layers such as layer 0 and layer 19 tend to produce generated videos that closely resemble the original when their attention maps are transferred. Interestingly, layer 3 stands out as an exception—its generation differs noticeably from the original but closely resembles the output from the source prompt. This suggests that layer 3 may play a key role in controlling the structural aspects of the generation. None of the other layers really have this characteristic.

Prior work has made use of cross-attention or text-attention transfer to enable editing~\citep{hertz2023prompttoprompt}. However, to the best of our knowledge, we are the first to show that fine-grained video editing is possible with simple self-attention transfer.

\subsection{The First Text Token is The Only Text Token That Matters}
\begin{figure}[!ht]
    \centering
    \renewcommand{\arraystretch}{0.5}
    \begin{tabular}{@{}m{0.10\textwidth} m{0.85\textwidth}@{}}
        \centering Original & \includegraphics[width=\linewidth]{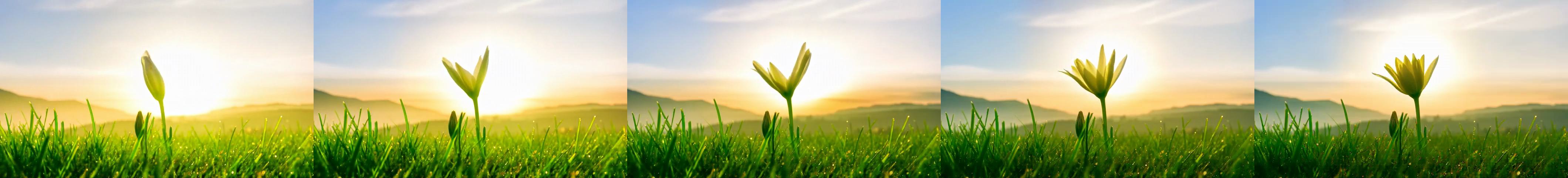} \\
        \centering Whole Model & \includegraphics[width=\linewidth]{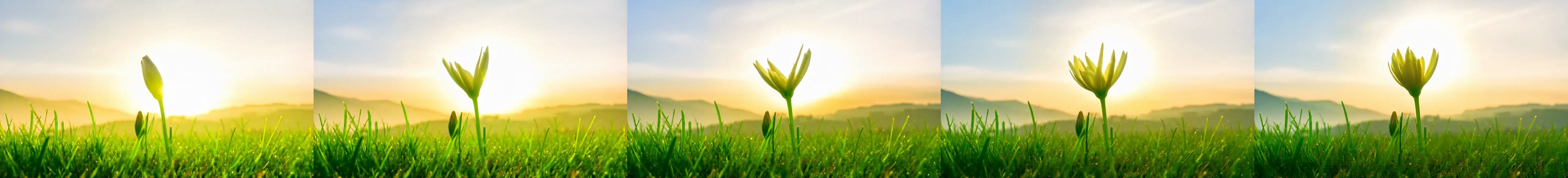} \\
    \end{tabular}
    \caption{\textbf{Generations with Only the First Text Token.} The prompt is: ``A time-lapse of a flower blooming in a vibrant meadow as the sun rises in the background.''}
    \label{fig:first_token}
\end{figure}

We also carefully inspect the model's attention to text tokens. \Cref{fig:attention_map} reveals that Mochi and HunyuanVideo both concentrate their attention on the first start-of-sequence text token \texttt{<s>}, even when processing long prompts. To understand why the model attends so heavily to it, we remove features of subsequent tokens. 
As shown in~\Cref{fig:first_token}, Mochi generates very similar videos when using only the features of \texttt{<s>}. Since the text encoder employs bi-directional attention, the first token may already capture a high-level summary of the prompt, making it sufficient for generation. Interestingly, at the second layer, we observe significant attention activity among text tokens attending to each other, but this appears to have a minimal impact on the generation.

\section{Attention Sparsity}
\label{sec:attn_sparsity}

Previous work on efficient \vdit{}s has revealed that their attention patterns are inherently sparse~\citep{katharopoulos2020transformers, zhang2025fast}. This insight has led to the development of approaches like replacing regular full attention with linear attention~\citep{ding2025efficient} and tiled attention~\citep{zhang2025fast}, which improve computational efficiency via sparsity. However, these methods often underperform without additional training to adapt to the newly introduced attention mechanisms.
In this section, we take a closer look at attention sparsity and provide new insights to better understand this limitation.

\subsection{Are They Really Sparse?}
\label{sec:are_they_sparse}

\begin{figure}[!ht]
    \centering
    \renewcommand{\arraystretch}{0.5}
    \begin{subfigure}[t]{\textwidth}
        \centering
        \begin{tabular}{@{}m{0.25\textwidth}@{}m{0.25\textwidth}@{}m{0.25\textwidth}@{}m{0.25\textwidth}@{}}
            \includegraphics[width=0.24\textwidth]{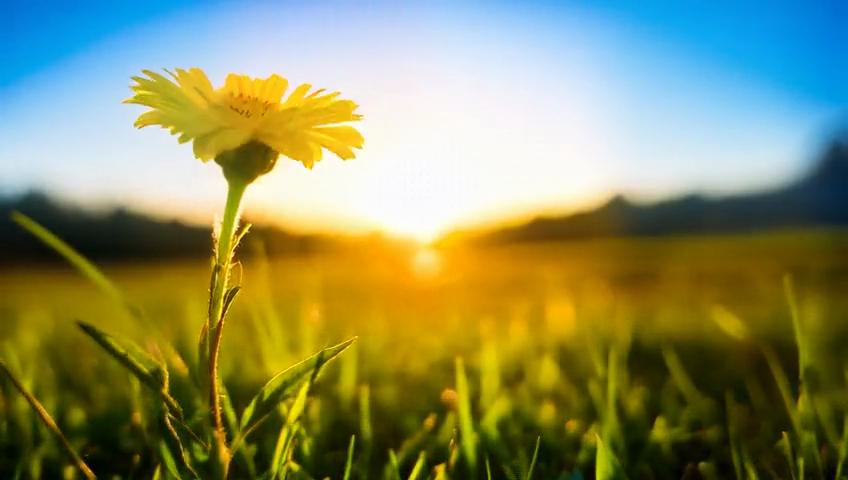} &\includegraphics[width=0.24\textwidth]{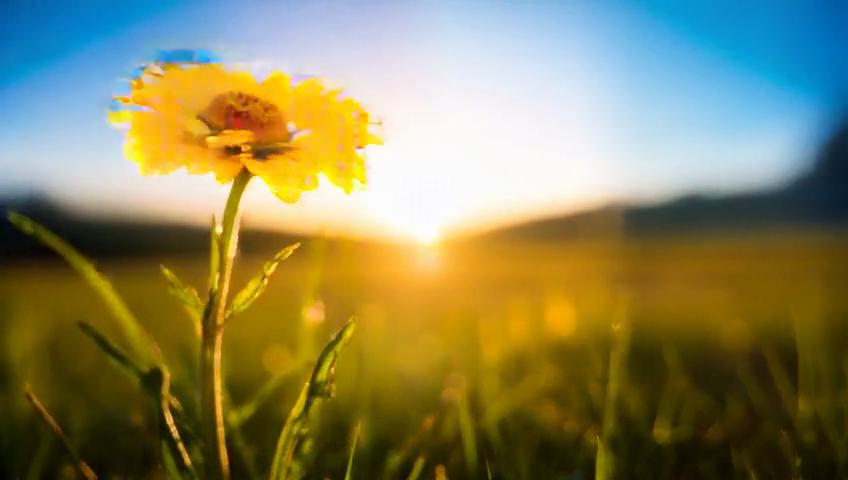} & \includegraphics[width=0.24\textwidth]{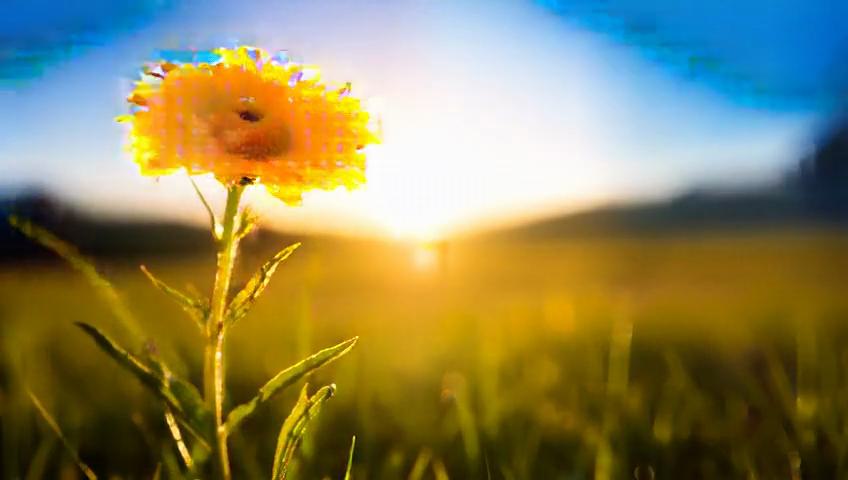}& \includegraphics[width=0.24\textwidth]{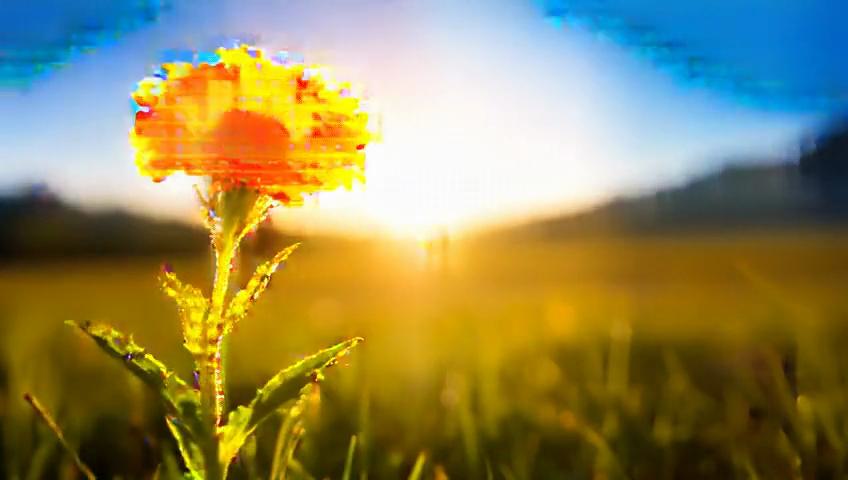} \\
            \multicolumn{1}{c}{\scriptsize Original} & \multicolumn{1}{c}{\scriptsize $10\%$} & \multicolumn{1}{c}{\scriptsize $15\%$} & \multicolumn{1}{c}{\scriptsize $20\%$} \\
        \end{tabular}
        \caption{\textbf{Masking out the Bottom $k\%$ Values in Attention Maps for All Layers.}}
        \label{fig:sparsity_figure_0}
    \end{subfigure}
    \hfill

    \begin{subfigure}[t]{\textwidth}
        \centering
        \begin{tabular}{@{}m{0.25\textwidth}@{}m{0.25\textwidth}@{}m{0.25\textwidth}@{}m{0.25\textwidth}@{}}
            \includegraphics[width=0.24\textwidth]{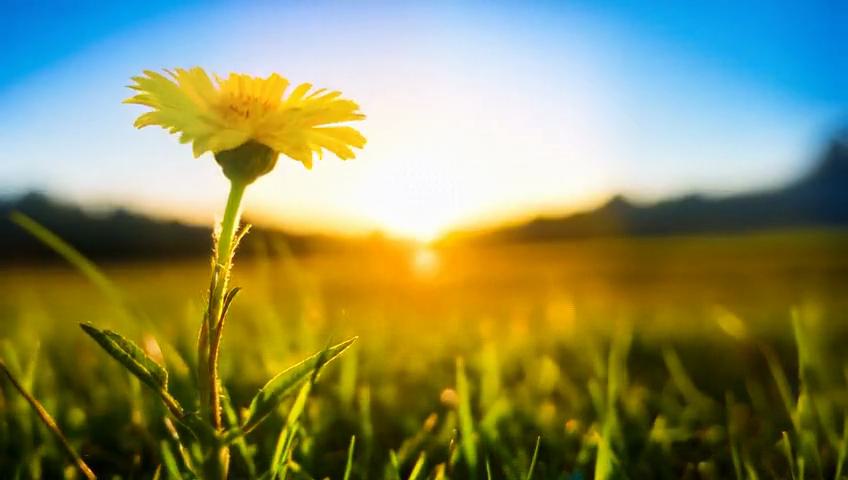} &\includegraphics[width=0.24\textwidth]{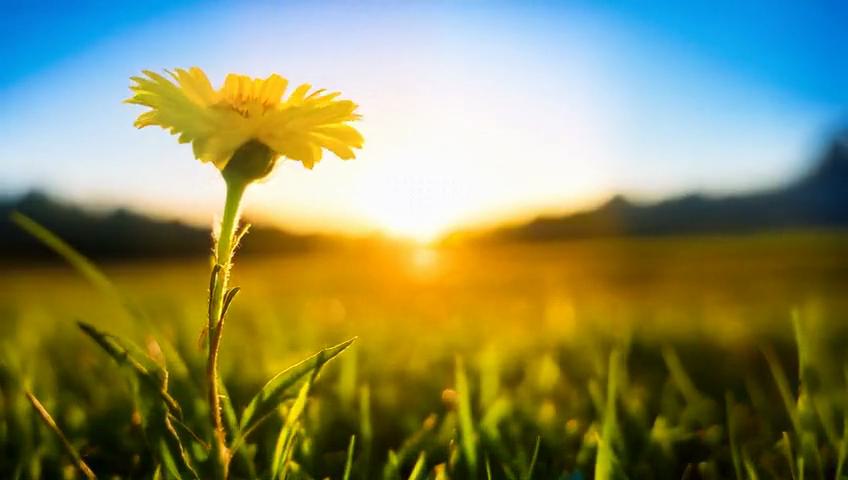} & \includegraphics[width=0.24\textwidth]{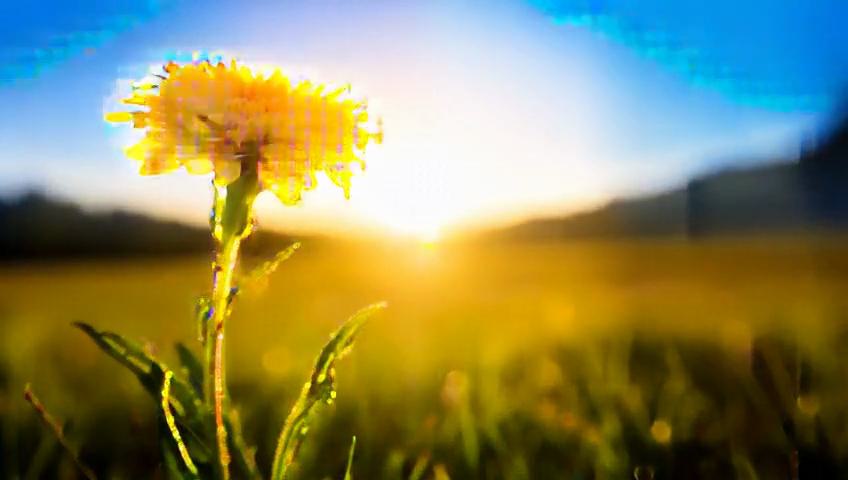}& \includegraphics[width=0.24\textwidth]{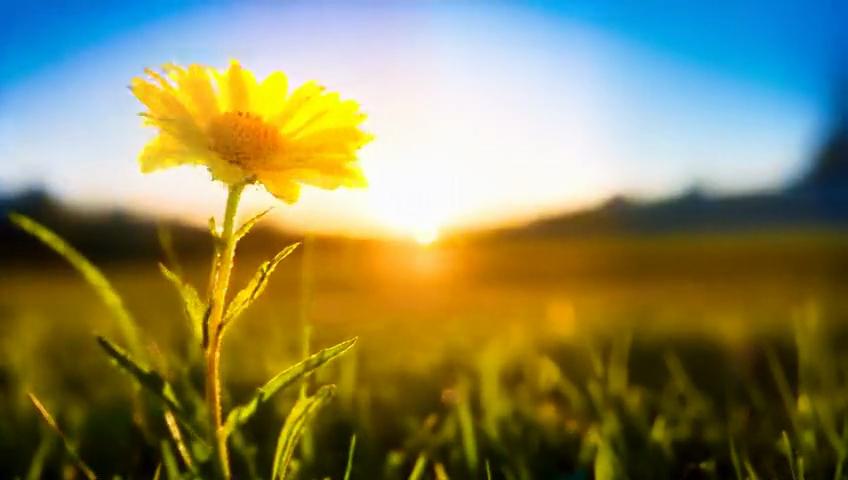} \\
            \multicolumn{1}{c}{\scriptsize Layer $5$} & \multicolumn{1}{c}{\scriptsize Layer $10$} & \multicolumn{1}{c}{\scriptsize Layer $44$} & \multicolumn{1}{c}{\scriptsize Layer $45$} \\
        \end{tabular}
        \caption{\textbf{Masking out the Bottom $20\%$ Values in Attention Maps for Only One Layer.}}
        \label{fig:sparsity_figure_1}
    \end{subfigure}
    \hfill

    \begin{subfigure}[t]{\textwidth}
        \centering
        \begin{tabular}{@{}m{0.25\textwidth}@{}m{0.25\textwidth}@{}m{0.25\textwidth}@{}m{0.25\textwidth}@{}}
            \includegraphics[width=0.24\textwidth]{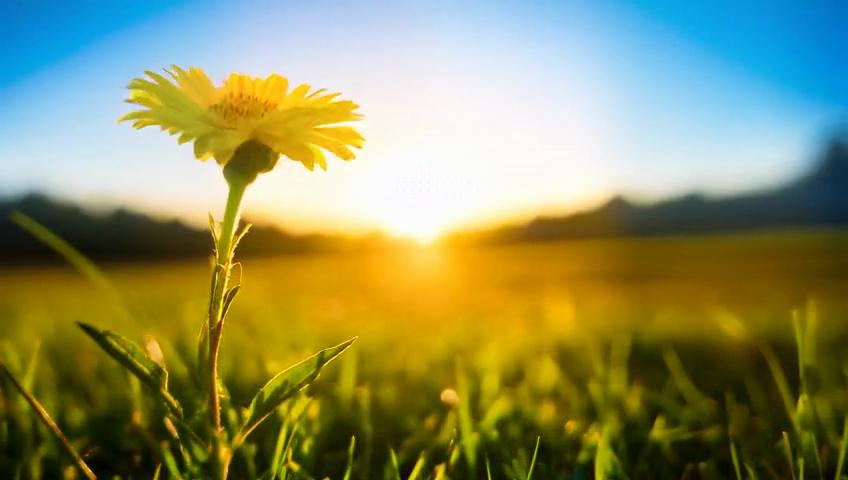} &\includegraphics[width=0.24\textwidth]{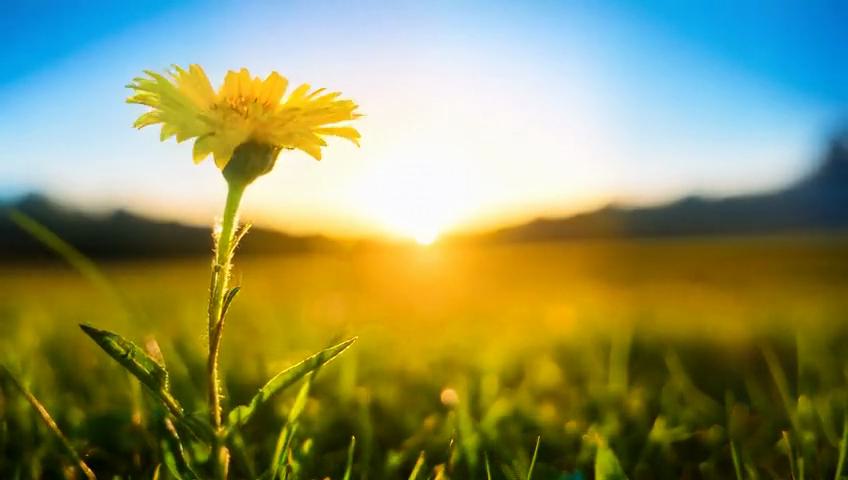} & \includegraphics[width=0.24\textwidth]{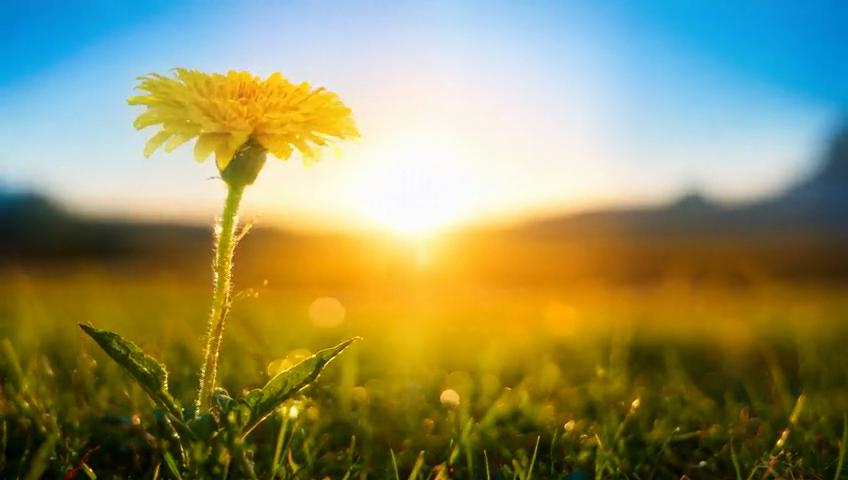}& \includegraphics[width=0.24\textwidth]{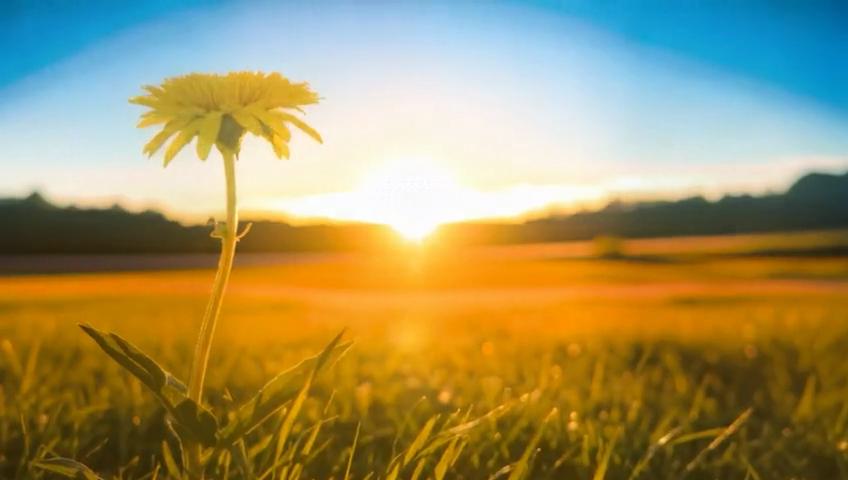} \\
            \multicolumn{1}{c}{\scriptsize $10\%$ Sparsity} & \multicolumn{1}{c}{\scriptsize $30\%$} & \multicolumn{1}{c}{\scriptsize $50\%$} & \multicolumn{1}{c}{\scriptsize $70\%$} \\
        \end{tabular}
        \caption{\textbf{Masking out the bottom $k\%$ Values in Attention Maps for All Layers except Layer $44$ and $45$.}}
        \label{fig:sparsity_figure_2}
    \end{subfigure}
    \hfill

    \caption{\textbf{Attention Sparsity Test.} The prompt is: ``A time-lapse of a flower blooming in a vibrant meadow as the sun rises in the background.'' Due to space limitations, additional frames are presented in Appendix~\Cref{fig:mask_bottom_k_all_layer}, \Cref{fig:mask_bottom_k_layer_wise}, and \Cref{fig:mask_bottom_k_except_two}.}
    \label{fig:sparsity_figure}
\end{figure}

Attention maps shown in~\Cref{fig:attention_map} appear sparse, suggesting that the attention mechanism could be optimized. 
To explore this, we apply the least destructive possible sparse attention implementation: masking out the bottom $k\%$ of Mochi's attention weights.
Specifically, for each denoising step, we run the model twice: first, with no sparsity to precompute attention weights and their $k\%$ quantile. Then, we rerun the model, masking out values below this threshold to achieve a $k\%$ mask ratio.
Masking is applied independently for each head and denoising step. Although computationally inefficient, this approach serves as an oracle proxy to explore the upper bound of achievable attention sparsity.

We present the results in~\Cref{fig:sparsity_figure_0}. Interestingly, even with $10\%$ sparsity, the generated videos exhibit noticeable quality degradation, including pixelation. This is counterintuitive, given our earlier observation that the attention maps appear highly sparse.

We investigate this by masking out one layer at a time to see if any specific layers are sensitive to such modifications. 
For most layers, the generation quality does not degrade at $k=20\%$. For example, layers $5$ and $10$ are robust, as seen in the left of~\Cref{fig:sparsity_figure_1}. However, there are two special layers that act quite differently. As shown in the right two examples of~\Cref{fig:sparsity_figure_1}, masking layers $44$ and $45$ noticeably degrades quality, even at merely $k=20\%$.

Based on the findings above, we further perform experiments by masking out the bottom $k\%$ of attention values across all layers \textit{except} for the two special layers: layers $44$ and $45$. As shown in~\Cref{fig:sparsity_figure_2}, the generations remain reasonably good even with a high mask rate of $70\%$, suggesting that the model’s attention is highly sparse across most layers.

To understand why these two layers stand out despite appearing similarly sparse, we visualized their bottom $k\%$ attention maps. However, we found no distinctive patterns—these maps closely resembled those of other layers. Further investigation into this phenomenon is a promising direction for future work.

\subsection{Temperature for \vdit{}s}
Temperature is a key hyperparameter in language transformers, often used to control output diversity and creativity. However, its role remains unexplored in the context of \vdit{}s. In our work, we apply temperature directly within the self-attention module to modulate attention sparsity—departing from the traditional approach, where temperature is applied solely to the output logits \citep{holtzman2019curious}. \\

\begin{figure}[!ht]
    \centering
    \renewcommand{\arraystretch}{0.5}
    \begin{tabular}{@{}m{0.10\textwidth} m{0.85\textwidth}@{}}
        \centering Original & \includegraphics[width=\linewidth]{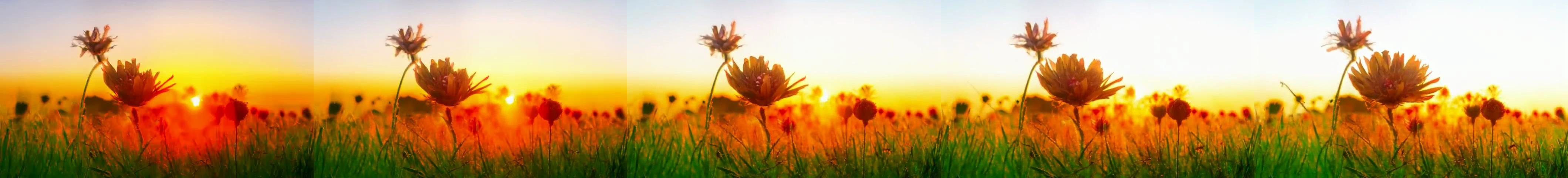} \\
        \centering $T=1.2$ & \includegraphics[width=\linewidth]{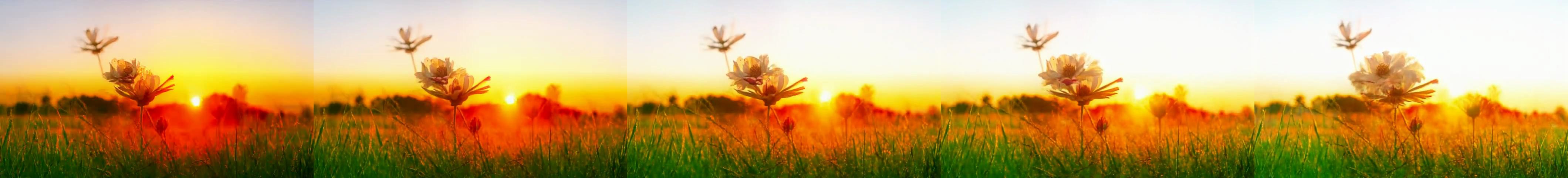} \\
        \centering $T=0.2$ & \includegraphics[width=\linewidth]{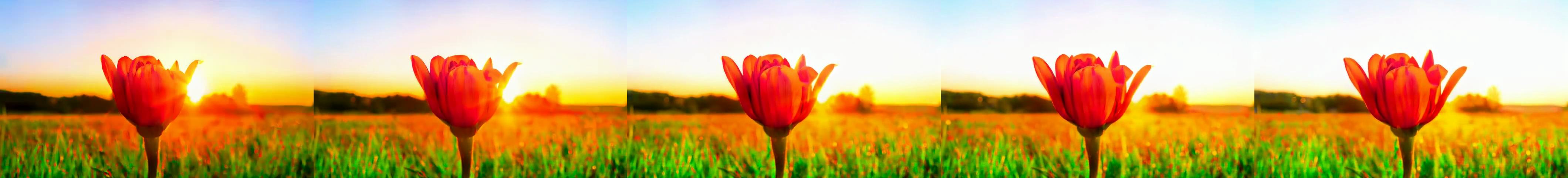} \\
    \end{tabular}
    \caption{\textbf{Generations with Temperature.} Modifying the temperature of just Layer $44$ results in dramatic changes in the generated video. Prompt: \textit{``A time-lapse of a flower blooming in a vibrant meadow as the sun rises in the background.''}}
    \label{fig:temperature}
\end{figure}

As shown in~\Cref{fig:temperature}, simply changing the temperature of one of Mochi's layers ($44$) results in significantly different generations. In general, we found that a temperature greater than~$1$ does not produce substantial improvements and sometimes leads to worse generations with many artifacts. However, using a lower temperature, such as the row with $T=0.2$, typically results in high-quality generations, occasionally even better than the original.

\section{Attention Sink}
\label{sec:attention_sink}
An attention sink \citep{xiao2023efficient} occurs when all queries in a transformer head attend disproportionately to a single key position, often yielding uninformative and atypical attention patterns. While well-studied in language models, this phenomenon has not, to our knowledge, been explored in \vdit{}s. Here, we show that attention sinks do arise in \vdit{}s, though with notable differences from their language model counterparts.

\subsection{Attention Sink in \vdit{}s}
\begin{figure}[htbp]
    \centering
    \begin{subfigure}[b]{0.35\textwidth}
        \centering
        \includegraphics[width=\textwidth]{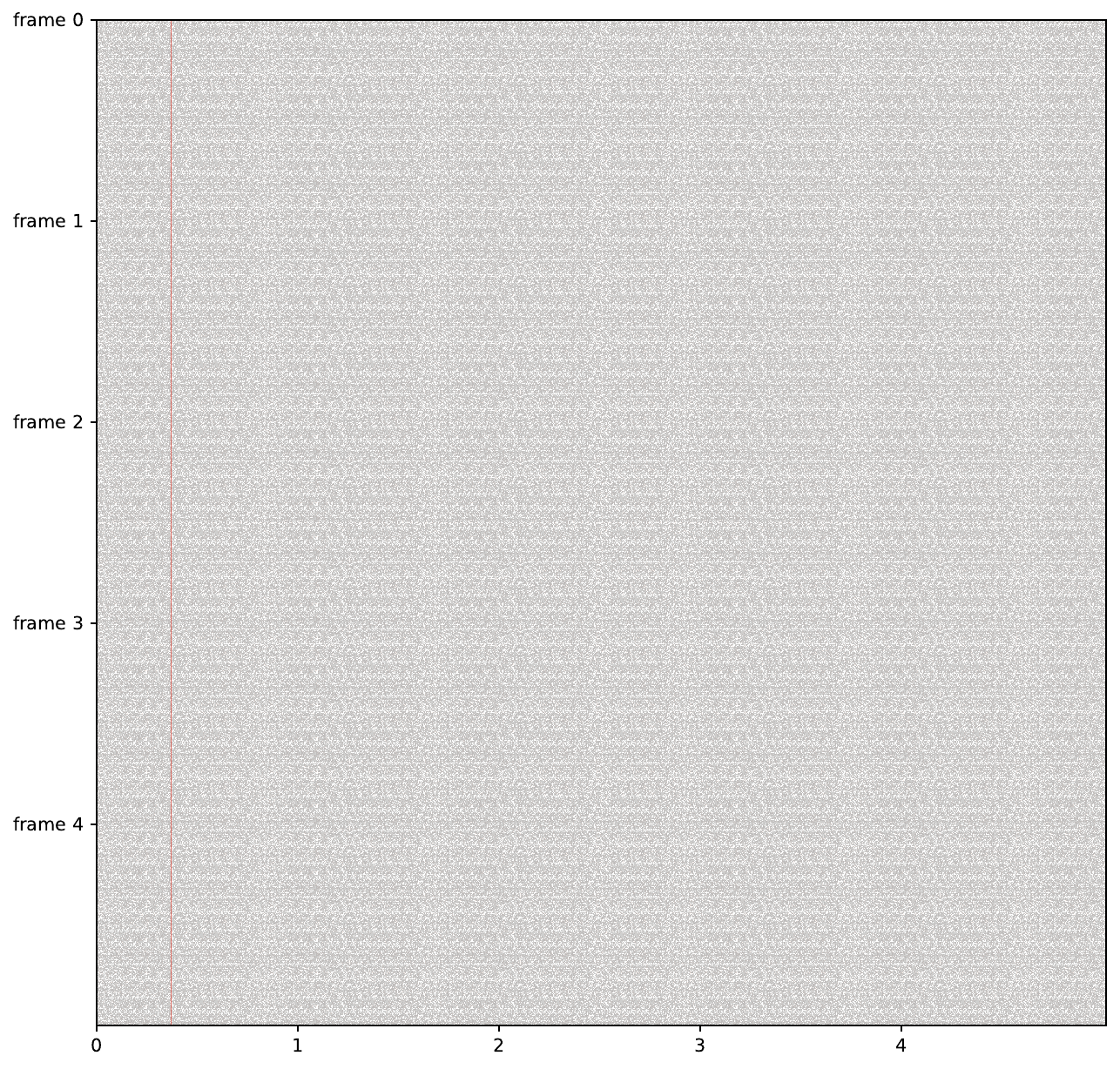}
        \caption{Attention Map of a Sink Head}
        \label{fig:attention_sink_0}
    \end{subfigure}
    \hspace{0.05\textwidth}
    \begin{subfigure}[b]{0.35\textwidth}
        \centering
        \includegraphics[width=\textwidth]{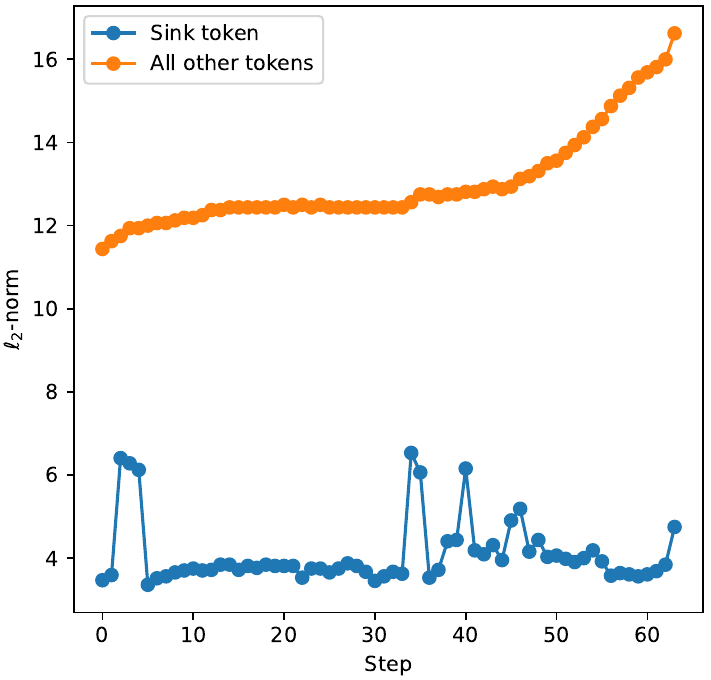}
        \caption{Value Norm}
        \label{fig:attention_sink_1}
    \end{subfigure}
    \caption{\textbf{Attention Sink in \vdit{}s.} In (b), the x-axis represents the denoising steps $0$ to $63$.}
    \label{fig:attention_sink}
\end{figure}

When visualizing attention in \vdit{}s, we found a striking pattern in some heads: a single vertical line in the attention maps (\Cref{fig:attention_sink_0}), resembling the attention sink seen in language models. This occurs consistently in Mochi, occasionally in Hunyuan, but not in Wan or CogVideoX. We focus on Mochi to explore the unique traits of attention sinks in \vdit{}s.

\begin{figure}[htbp]
    \centering
    \renewcommand{\arraystretch}{0.5}
    \begin{tabular}{@{}m{0.10\textwidth} m{0.85\textwidth}@{}}
        \centering Original & \includegraphics[width=\linewidth]{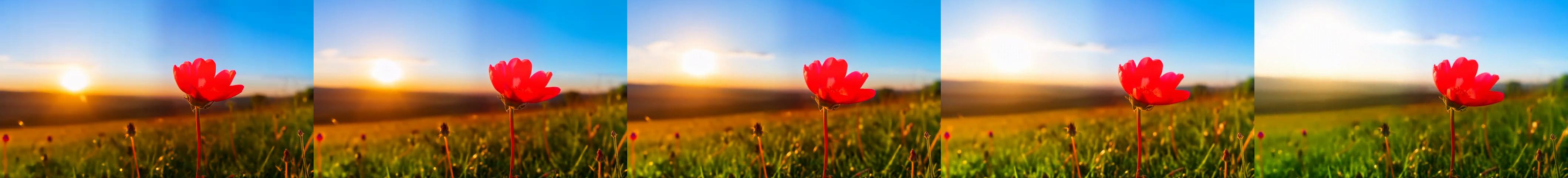} \\
        \centering Skip Sink Head & \includegraphics[width=\linewidth]{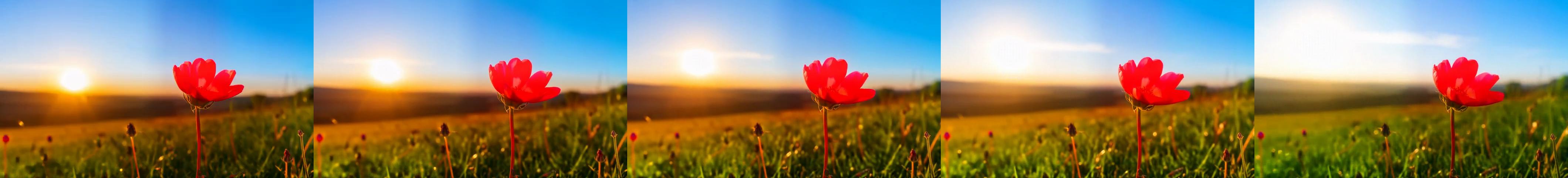} \\
        \centering Skip Random Head & \includegraphics[width=\linewidth]{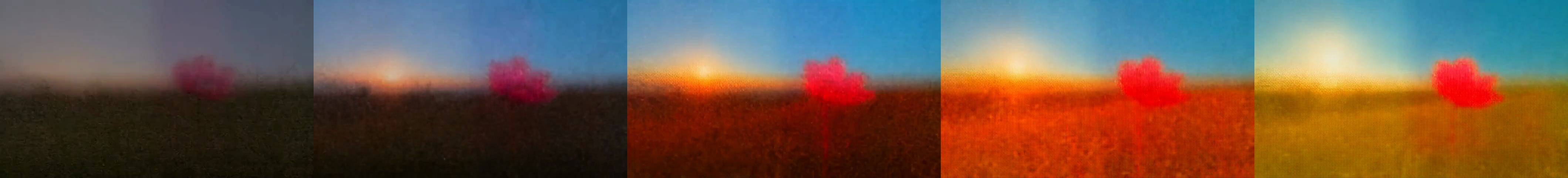} \\
    \end{tabular}
    \caption{\textbf{Skipping Attention Sink Heads.} Prompt: \textit{``A time-lapse of a flower blooming in a vibrant meadow as the sun rises in the background.''} For the random head skipping scenario, we randomly skip the same number of attention heads as in the attention sink head case. In both scenarios, approximately $4\%$ of the total heads are skipped.}
    \label{fig:sink_skip}
\end{figure}

\paragraph{Is It Attention Sink?}
We first ask whether the vertical line pattern in \vdit{}s reflects the attention sink phenomenon in language models. In LMs, attention sinks typically contribute little: all queries attend to the first token (often a special token like \texttt{<BOS>}), which carries minimal information due to the causal mask \citep{gu2024attention}. These sinks are thus seen as no-ops. To determine if the same holds in \vdit{}s, we examine whether these vertical patterns are similarly uninformative.

We compare the value norms of Mochi's sink tokens to the average across other tokens in~\Cref{fig:attention_sink_1}. Sink tokens have significantly smaller norms, suggesting minimal contribution to final representations. This mirrors language models, where low-value norms lead to near no-op behavior when all queries attend to the sink \citep{gu2024attention}.

\paragraph{Similar Property to LLMs: Small Impact to Generation}
To test their impact, we skip attention sink heads during generation. A head is marked as a sink if over half the tokens heavily attend to a single position; we then zero its output. As shown in~\Cref{fig:sink_skip}, skipping all sink heads preserves output quality, while skipping the same number of random non-sink heads degrades it. This suggests that sink heads contribute little and are largely redundant.

\subsection{Unique Characteristics}
In this subsection, we examine the patterns of attention sinks in \vdit{}s, which differ significantly from those observed in language models.
\begin{wrapfigure}{r}{0.4\textwidth}
    \centering
    \includegraphics[width=\linewidth]{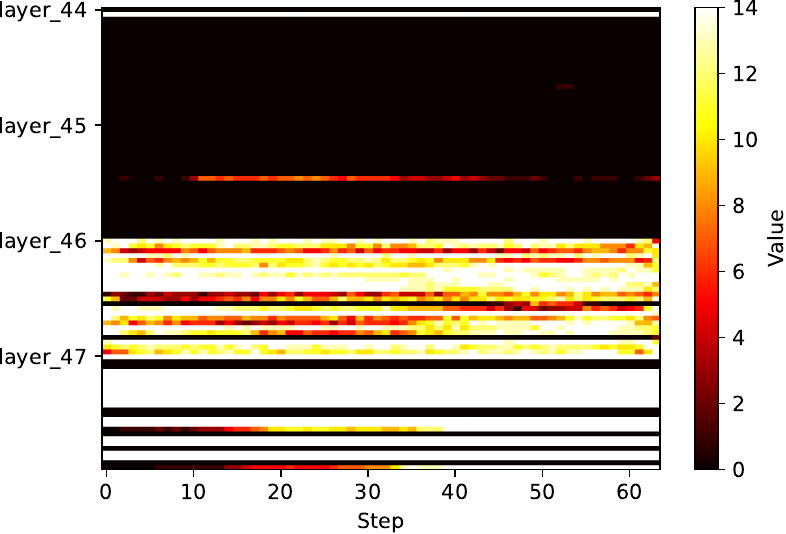}
    \vspace{-0.5cm}
    \caption{\textbf{Frequency of Attention Sink per Head.}}
    \vspace{-0.5cm}
    \label{fig:attention_consistency}
\end{wrapfigure}

\textbf{Layers and Heads Consistency.} While attention sinks in language models can appear in any layer \citep{gu2024attention}, in \vdit{}s, they predominantly emerge in later layers. In Mochi, sinks appear only in the last four layers—most frequently in the last two. These sinks are highly consistent, occurring in the same heads across different prompts and inference settings. \Cref{fig:attention_consistency} shows sink frequency across 14 prompts, with each row representing a head and the x-axis indicating the denoising step; higher values indicate more frequent sink behavior. \Cref{fig:attention_consistency} reveals that layers $44$ and $45$ have few sink heads (typically one), with only one head in layer $44$ showing consistent behavior. In contrast, over $80\%$ of heads in layers $46$ and $47$ consistently exhibit sink patterns.

\begin{figure}[htbp]
    \centering
    \begin{subfigure}[b]{0.4\textwidth}
        \centering
        \includegraphics[width=\textwidth]{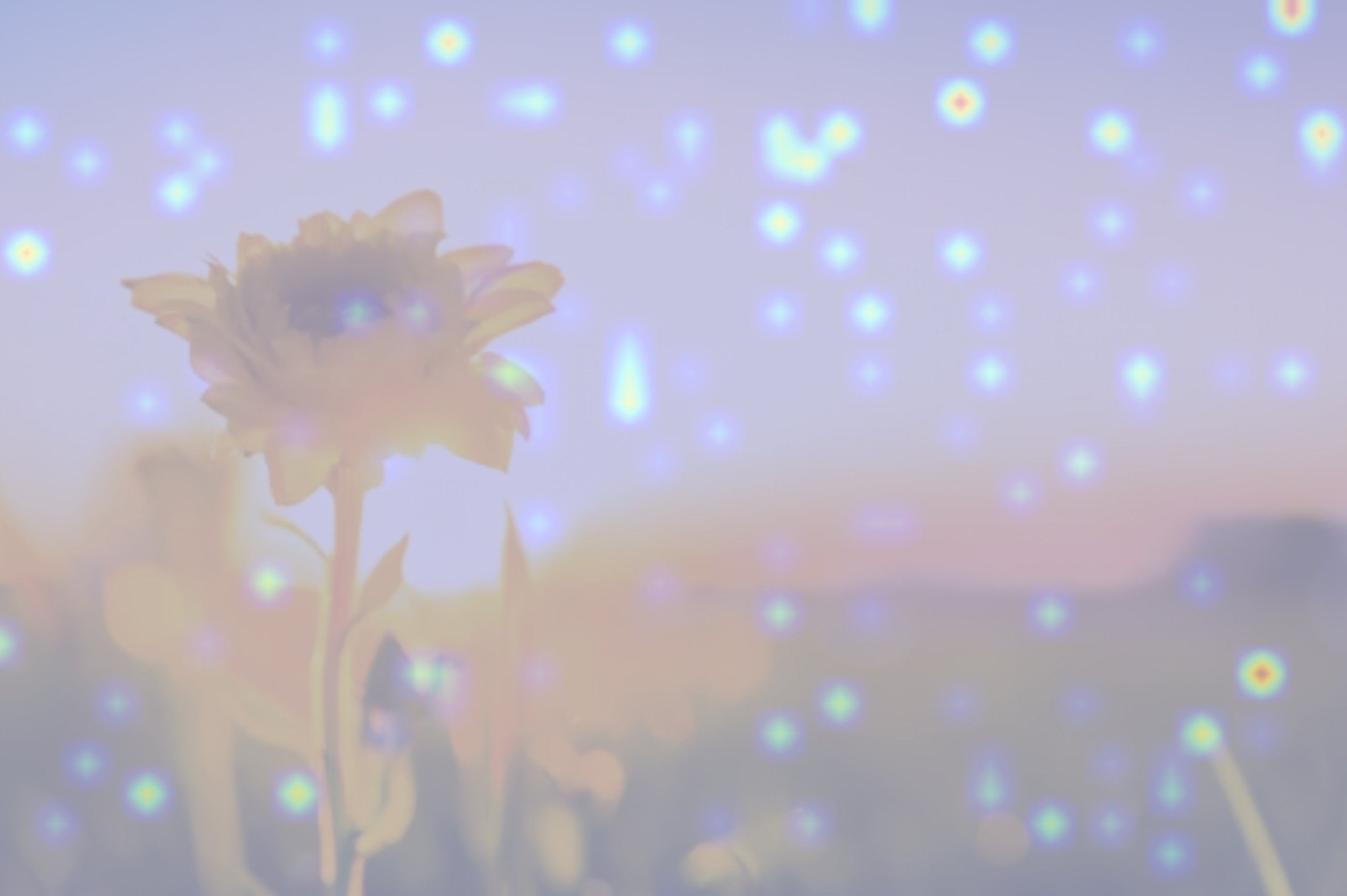}
        \caption{Spatial Distribution of Sink Tokens}
        \label{fig:attention_inconsistency_0}
    \end{subfigure}
    \hspace{0.05\textwidth}
    \begin{subfigure}[b]{0.4\textwidth}
        \centering
        \includegraphics[width=0.9\textwidth]{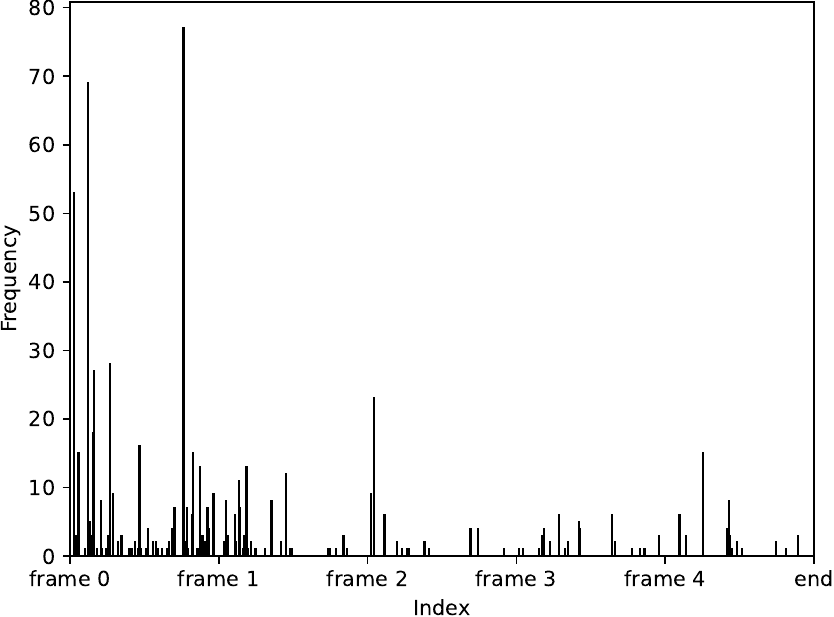}
        \caption{Temporal Distribution of Sink Tokens}
        \label{fig:attention_inconsistency_1}
    \end{subfigure}
    \caption{\textbf{Distribution of Sink Tokens.}}
    \label{fig:attention_inconsistency}
\end{figure}

\textbf{Spatial Randomness.} As shown in the heatmap in~\Cref{fig:attention_inconsistency_0}, sink tokens are randomly scattered across the spatial space, without consistent clustering around important or unimportant regions. Their locations vary widely across prompts, showing no clear pattern.

\textbf{Temporal Bias.} As shown in~\Cref{fig:attention_inconsistency_1}, sink tokens exhibit a clear temporal bias—appearing mostly in the first latent frame and decreasing in later ones. This contrasts with the spatial distribution, which shows no consistent pattern.

\section{Increasing Sparsity and Removing Sinks through Retraining}
\label{sec:retraining}

Past work links the emergence of attention degeneracies like sinks to optimization dynamics~\citep{gu2024attention}. To improve our ability to sparsify \vdit{}s, we propose a simple yet effective mitigation strategy: retraining problematic layers from scratch.

We start with a variant of the Mochi model which also exhibits issues with sparsity in layers $44$ and $45$ (\Cref{sec:attn_sparsity}) and sinks in layers $46$ and $47$ (\Cref{sec:attention_sink}) after pre-training. Similar to the original Mochi, masking attention in these layers even $20\%$ severely degrades quality.

We reinitialize Mochi's final four transformer blocks ($44$-$47$) with standard parameterization and retrain them on a large captioned video dataset. The rest of the model is frozen, greatly accelerating training. We use AdamW with a constant learning rate, a short warmup, and EMA tracking, following standard diffusion model pretraining practices.

\begin{figure}[!t]
\centering
    \renewcommand{\arraystretch}{0.5}
    \begin{tabular}{@{}m{0.10\textwidth}@{}m{0.75\textwidth}@{}m{0.12\textwidth}@{}}
    \centering \small Before & \includegraphics[width=\linewidth]{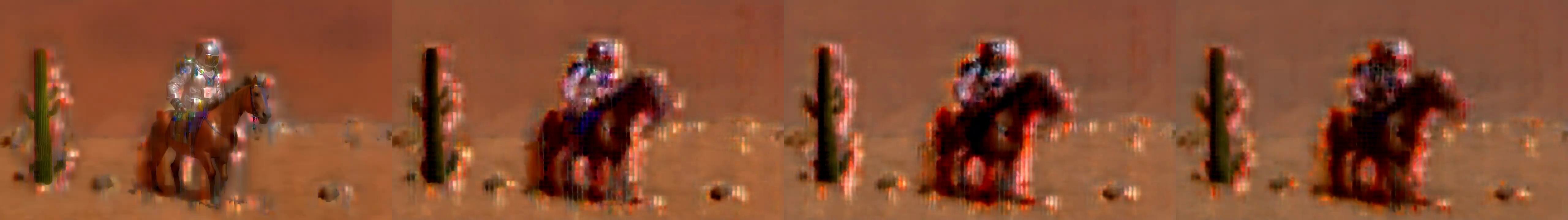} & \includegraphics[width=\linewidth]{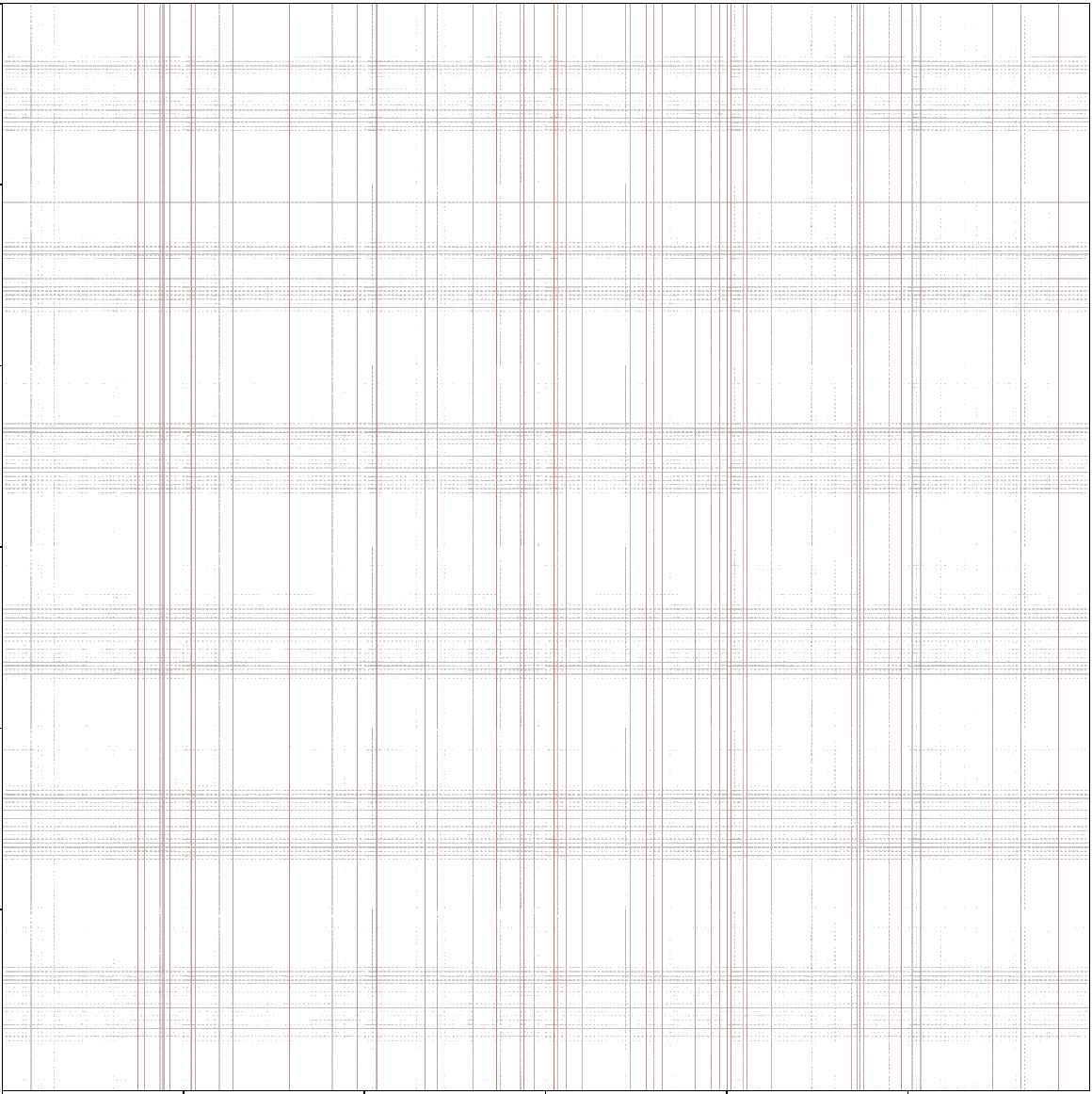} \\
    \centering \small After & \includegraphics[width=\linewidth]{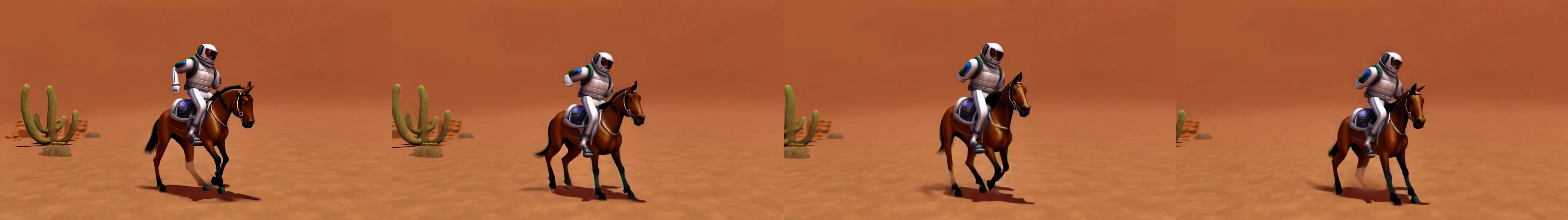} & \includegraphics[width=\linewidth]{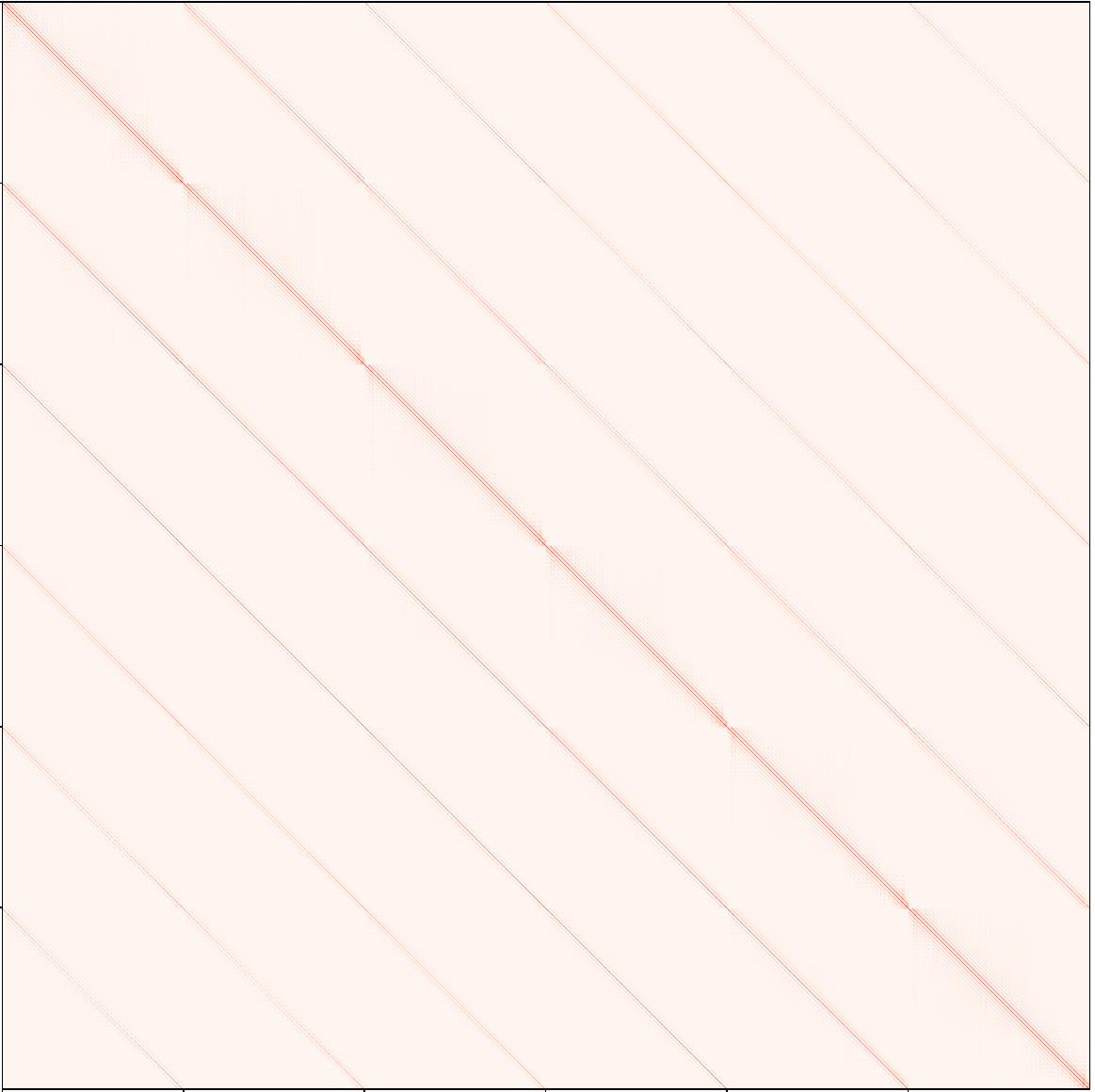} \\
     & \centering \small Generation with Bottom $20\%$ Masking & \centering \small Attention \\
    \end{tabular}
    \caption{\textbf{Increasing Sparsity and Removing Sinks through Retraining.} Prompt: \textit{``An astronaut riding a horse on Mars with a cactus in the background.''}}
    \label{fig:attn_reinit}
\end{figure}

After retraining, the final four layers show sparse attention patterns consistent with the rest of the network—dominant diagonal and off-diagonal attention, with no visible sinks~(\Cref{fig:attn_reinit}). Now, sparsifying the bottom $20\%$ of attention values has no impact on output quality, suggesting improved efficiency and better use of model capacity.

\section{Insights and Future Directions}
Based on our analysis, we highlight several promising directions:

\textbf{Efficient Attention.} Despite varying architectures and training data, models converge to similar attention patterns. Initializing models with these patterns or using designed masks could improve \vdit{} efficiency.

\textbf{Temperature Tuning.} Adjusting the temperature of a single layer can significantly affect generation, suggesting it as a useful control knob. Learnable temperature parameters may also aid in fine-tuning and personalization.

\textbf{Layer-Wise Sparsity Matters.} Our results suggest that not all layers contribute equally to sparsity, and thus, special care should be paid when deciding which layers to prune or sparsify. A more nuanced, layer-wise analysis is necessary before applying uniform sparsification strategies.

\textbf{Video Editing via Attention Transfer.} Self-attention transfer offers a path to targeted video editing. Since specific layers influence distinct aspects (e.g., camera angle), modifying a targeted subset of layers may allow efficient, fine-grained control.

\textbf{Fine-grained Text Control.} While focusing on the first text token can improve efficiency, it may hinder fine-grained generation by ignoring the rest of the prompt. To enable more precise control, models may need to distribute attention more evenly across all tokens.

\section{Conclusion}
In this work, we provided the first comprehensive analysis of attention mechanisms in Video Diffusion Transformers (\vdit{}s), uncovering three key properties—Samey, Sparse, and Sinky. We demonstrated how attention patterns remain consistent across prompts and models, enabling novel video editing via attention transfer. Our sparsity analysis revealed that not all seemingly sparse layers can be effectively sparsified without quality loss, while certain layers are particularly sensitive. We also identified and characterized attention sinks in \vdit{}s, showing their minimal contribution and proposing retraining as a mitigation strategy. These findings offer practical insights and open new directions for improving the efficiency, interpretability, and controllability of video diffusion models.

\section{Acknowledgements}
This material is based upon work partially supported by the NSF Grant No. 2229885 (NSF Institute for Trustworthy AI in Law and Society, TRAILS). Any opinions, findings and conclusions or recommendations expressed in this material are those of the author(s) and do not necessarily reflect the views of the National Science Foundation.

\bibliography{colm2025_conference}
\bibliographystyle{colm2025_conference}

\appendix
\section{Appendix}
\begin{figure}[!ht]
    \centering
    \renewcommand{\arraystretch}{0.5}
    \begin{tabular}{@{}m{0.10\textwidth} m{0.85\textwidth}@{}}
        \centering Original & \includegraphics[width=\linewidth]{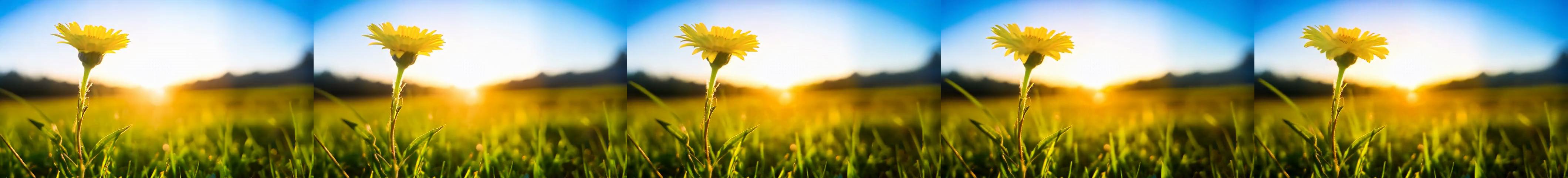} \\
        \centering $10\%$ Sparsity & \includegraphics[width=\linewidth]{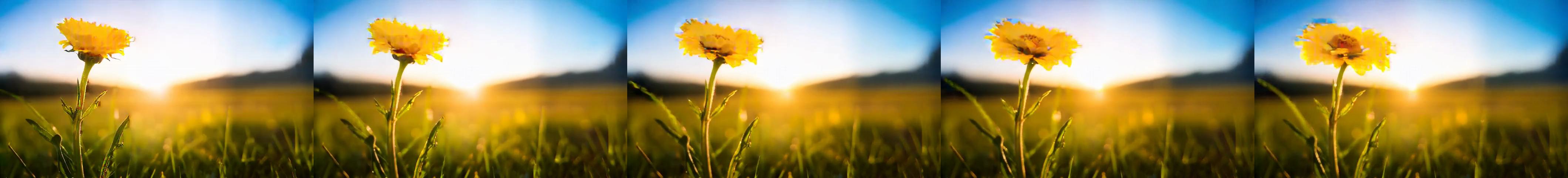} \\
        \centering $15\%$ & \includegraphics[width=\linewidth]{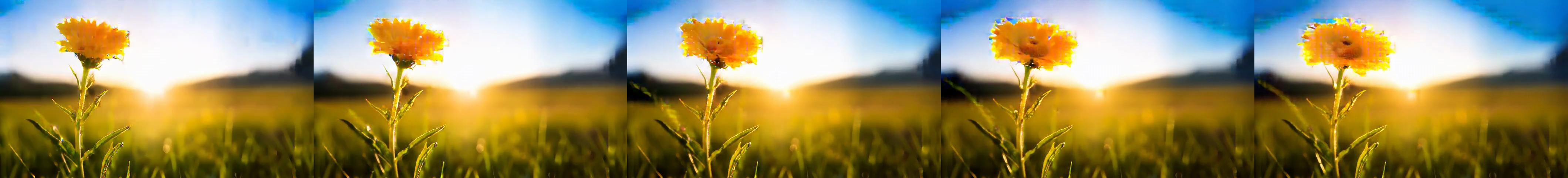} \\
        \centering $20\%$ & \includegraphics[width=\linewidth]{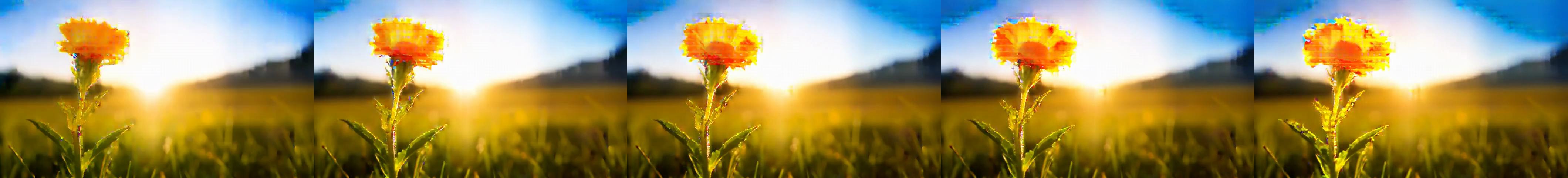} \\
    \end{tabular}
    \caption{\textbf{Masking out the Bottom $k\%$ Values in Attention Maps for All Layers.} The prompt is: ``A time-lapse of a flower blooming in a vibrant meadow as the sun rises in the background.''}
    \label{fig:mask_bottom_k_all_layer}
\end{figure}

\begin{figure}[!ht]
    \centering
    \renewcommand{\arraystretch}{0.5}
    \begin{tabular}{@{}m{0.10\textwidth} m{0.85\textwidth}@{}}
        \centering Layer $5$ & \includegraphics[width=\linewidth]{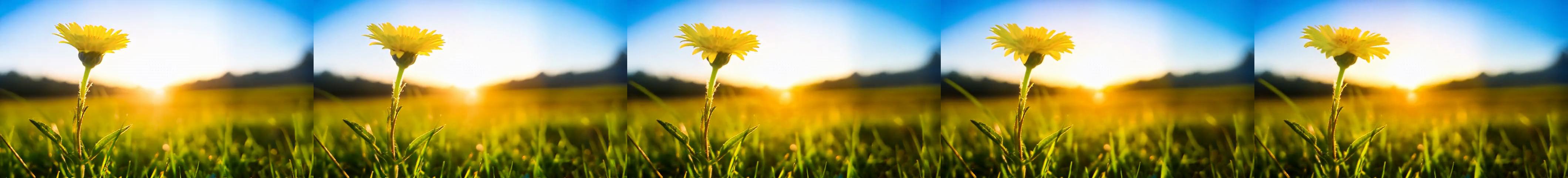} \\
        \centering Layer $10$ & \includegraphics[width=\linewidth]{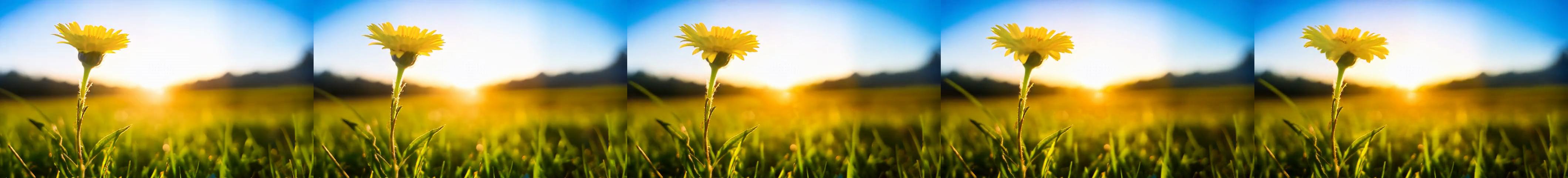} \\
        \centering Layer $44$ & \includegraphics[width=\linewidth]{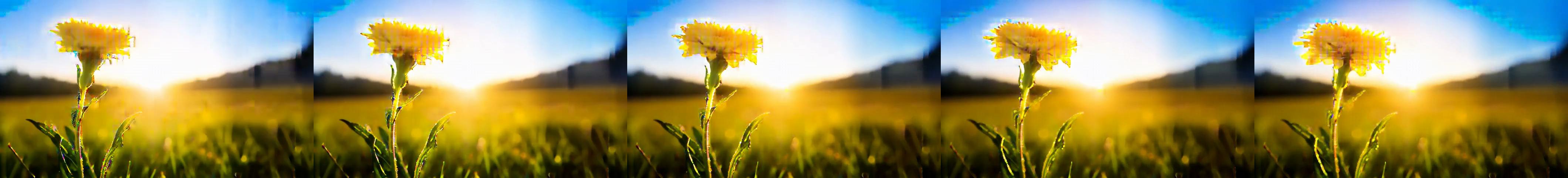} \\
        \centering Layer $45$ & \includegraphics[width=\linewidth]{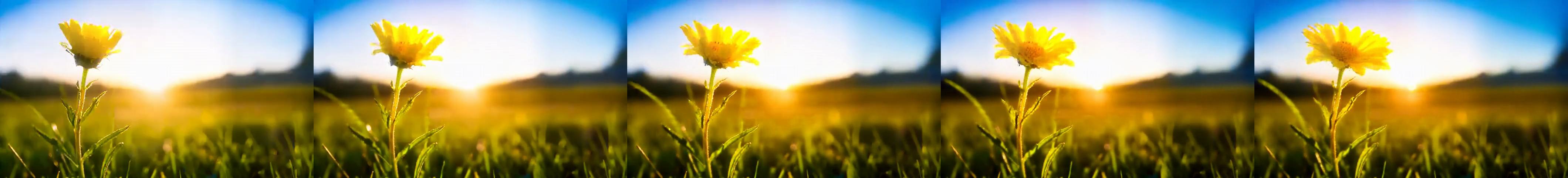} \\
    \end{tabular}
    \caption{\textbf{Masking out the Bottom $20\%$ Values in Attention Maps for Only One Layer.} The prompt is: ``A time-lapse of a flower blooming in a vibrant meadow as the sun rises in the background.''}
    \label{fig:mask_bottom_k_layer_wise}
\end{figure}

\begin{figure}[!ht]
    \centering
    \renewcommand{\arraystretch}{0.5}
    \begin{tabular}{@{}m{0.10\textwidth} m{0.85\textwidth}@{}}
        \centering $10\%$ Sparsity & \includegraphics[width=\linewidth]{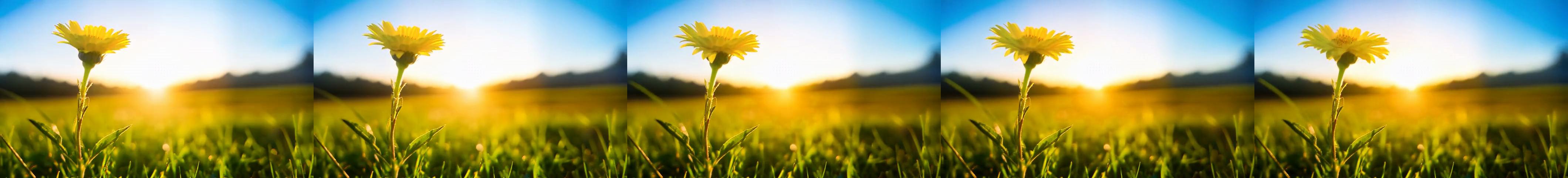} \\
        \centering $30\%$ & \includegraphics[width=\linewidth]{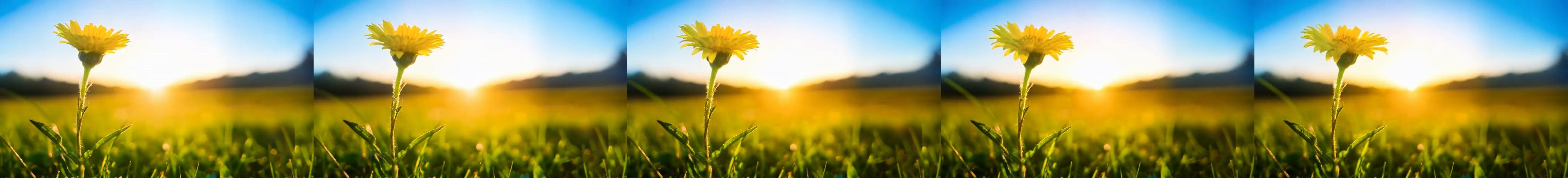} \\
        \centering $50\%$ & \includegraphics[width=\linewidth]{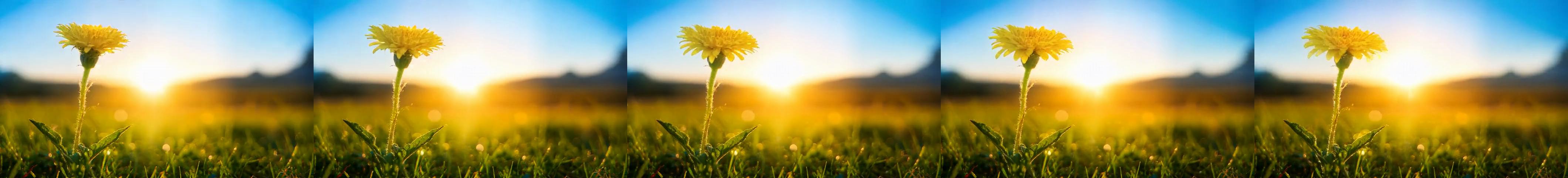} \\
        \centering $70\%$ & \includegraphics[width=\linewidth]{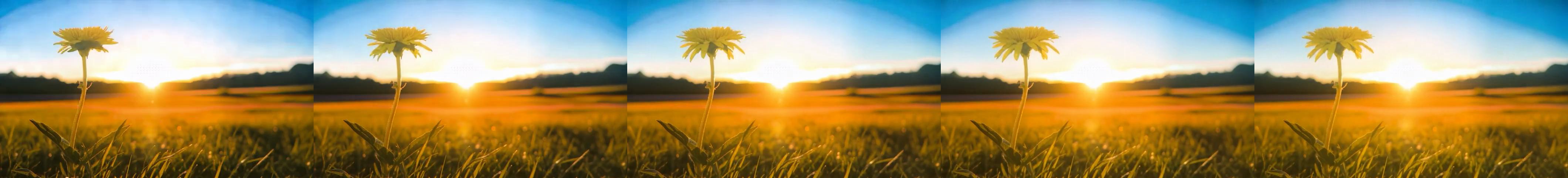} \\
    \end{tabular}
    \caption{\textbf{Masking out the bottom $k\%$ Values in Attention Maps for All Layers except Layer $44$ and Layer $45$.} The prompt is: ``A time-lapse of a flower blooming in a vibrant meadow as the sun rises in the background.''}
    \label{fig:mask_bottom_k_except_two}
\end{figure}

\end{document}